\documentclass[runningheads]{llncs}

\usepackage[T1]{fontenc}
\usepackage{graphicx}
\usepackage{rotating}

\begin{document}

\title{Semantic Segmentation using Neural Ordinary Differential Equations}

\titlerunning{Semantic Segmentation using Neural ODEs}

\author{
Seyedalireza Khoshsirat\orcidID{0000-0002-6659-0707}
\and Chandra Kambhamettu\orcidID{0000-0001-5306-3994}
}

\authorrunning{S. Khoshsirat, C. Kambhamettu}

\institute{VIMS Lab, University of Delaware, Newark DE 19716, USA\\
\email{\{alireza,chandrak\}@udel.edu}}

\maketitle              

\begin{abstract}
The idea of neural Ordinary Differential Equations (ODE) is to approximate the derivative of a function (data model) instead of the function itself.
In residual networks, instead of having a discrete sequence of hidden layers, the derivative of the continuous dynamics of hidden state can be parameterized by an ODE.
It has been shown that this type of neural network is able to produce the same results as an equivalent residual network for image classification.
In this paper, we design a novel neural ODE for the semantic segmentation task.
We start by a baseline network that consists of residual modules, then we use the modules to build our neural ODE network.
We show that our neural ODE is able to achieve the state-of-the-art results using 57\% less memory for training, 42\% less memory for testing, and 68\% less number of parameters.
We evaluate our model on the Cityscapes, CamVid, LIP, and PASCAL-Context datasets.

\end{abstract}

\section{Introduction}
\textbf{Neural Ordinary Differential Equations.}
In machine learning, we try to iteratively find a function that best describes the data.
There are two basic approaches to finding this function.
The first approach is to directly approximate the function by an analytical or numerical method.
An ordinary linear regression falls into this category.
The second approach is to approximate the derivative of the function.
This results in an Ordinary Differential Equation (ODE) which by solving, we get the approximation of the function.
We can parameterize the derivative of the function as a neural network.
\par
Now, consider a residual network \cite{he2016identity} where all the hidden states have the same dimension.
Such networks generate an output by doing a sequence of transformations to a hidden state \cite{chen2018neural}:
\begin{equation} \label{eq:1}
 \mathbf{h}_{t+1} = \mathbf{h}_t + f(\mathbf{h}_t, \theta_t)
\end{equation}
By adding infinite number of layers, we get the continuous dynamics of hidden units using an ODE defined by a neural network \cite{chen2018neural}:
\begin{equation}
 \frac{d\mathbf{h}(t)}{dt} = f(\mathbf{h}(t), t, \theta)
\end{equation}
where $f(\mathbf{h}(t), t, \theta)$ is a neural network layer parameterized by $\theta$ at layer $t$.
By solving the integral:
\begin{equation}
 \mathbf{h}(t) = \mathbf{h}(t_0) + \int_{t_0}^{T} f(\mathbf{h}(t), t, \theta) \, dt,
\end{equation}
we can get the output value of a hidden layer at some depth $T$.
\par

\textbf{Semantic Segmentation.}
Semantic segmentation refers to the process of assigning each pixel in an image to a class label.
Current state-of-the-art neural networks for semantic segmentation require a considerable amount of memory for training (especially with high-resolution images).
Based on the fact that neural ODEs use less memory \cite{chen2018neural}, in this paper we propose a novel neural ODE design for the semantic segmentation task.
We evaluate our model on the Cityscapes \cite{cordts2016cityscapes}, CamVid \cite{BrostowFC:PRL2008}, LIP \cite{gong2017look}, and PASCAL-Context \cite{mottaghi2014role} datasets and show that it is able to produce the state-of-the-art results using 57\% less memory for training, 42\% less memory for testing, and 68\% less number of parameters.

\begin{figure}[t]
\includegraphics[width=\textwidth]{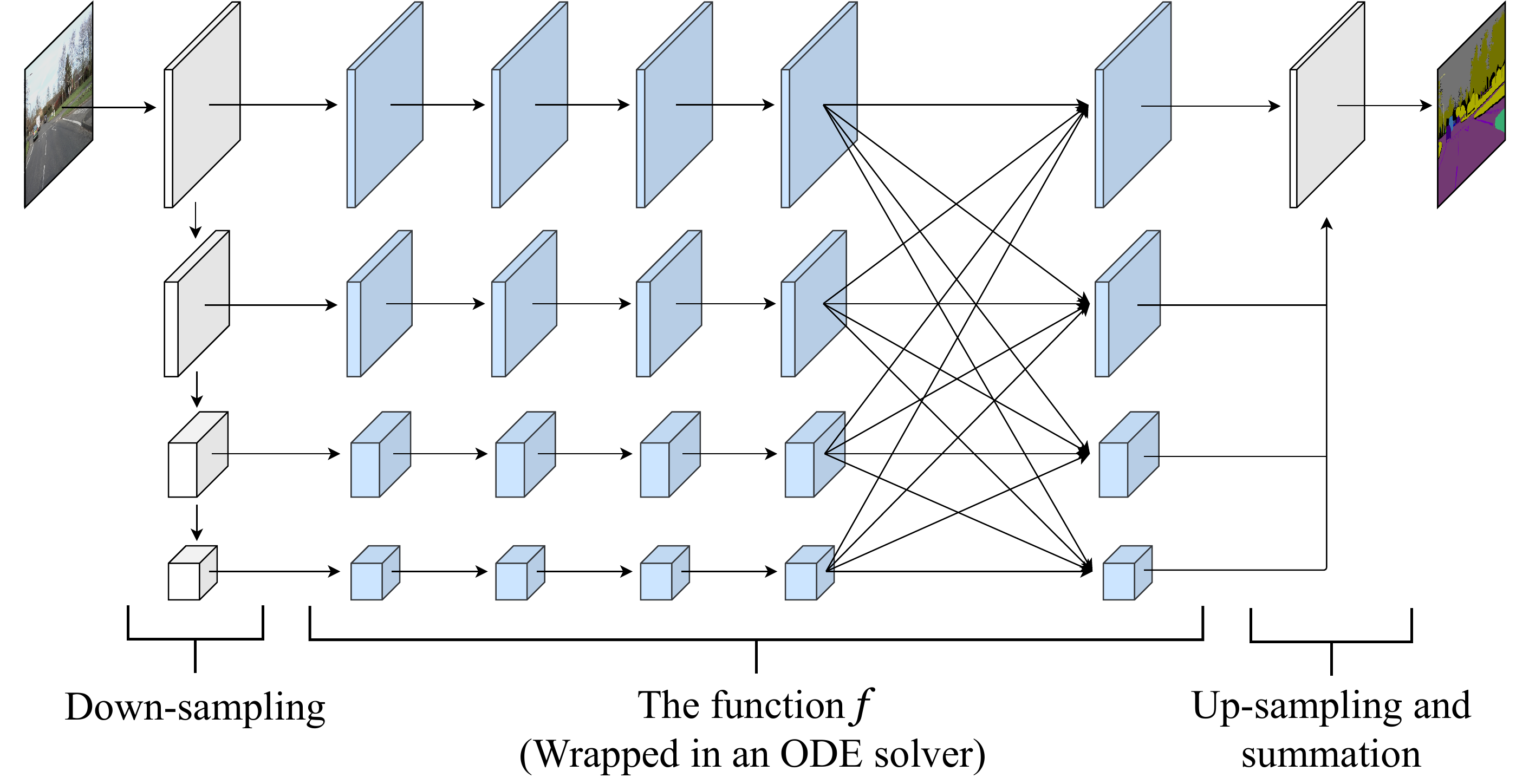}
\caption{The overall structure of the proposed method.
The down-sampling uses convolutions with $stride=2$.
For the up-sampling, we use bilinear interpolation to avoid the checkerboard artifact \cite{odena2016deconvolution}.}
\label{fig:segnode}
\end{figure}

\section{Related Work}
{\bf Neural ODEs.}
Recently, several works have analyzed the relationship between dynamical systems and deep neural networks.
In \cite{weinan2017proposal}, the authors propose the idea of using continuous dynamical systems as a tool for machine learning.
In \cite{lu2017beyond}, it has been shown that many effective networks, such as ResNet \cite{he2016identity}, PolyNet \cite{zhang2017polynet}, FractalNet \cite{larsson2016fractalnet}, and RevNet \cite{gomez2017reversible}, can be interpreted as different numerical discretizations of differential equations.
When the discretization step approaches zero, it yields a family of neural networks, which are called neural ODEs \cite{chen2018neural}.
\cite{chen2018neural} proposes to compute gradients using the adjoint sensitivity method \cite{pontryagin2018mathematical}, in which there is no need to store intermediate quantities during the forward pass of the network.
In \cite{zhu2018convolutional}, an interpretation of Dense Convolutional Networks (DenseNets) \cite{huang2017densely} and Convolutional Neural Networks with Alternately Updated Clique (CliqueNets) \cite{yang2018convolutional} is provided from a dynamical systems view point.
\par
{\bf Memory Usage Reduction.}
There are methods to reduce memory footprints.
Reduced precision formats are binary floating-point formats that occupy less than 32 bits (four bytes) \cite{courbariaux2014low,micikevicius2017mixed}.
These formats either reduce the accuracy or add some processing overhead for converting high precision to low precision.
Many other memory reduction techniques are derivatives of binomial gradient check-pointing \cite{griewank2000algorithm,chen2016training,rota2018place}.
The overall idea of gradient checkpointing is that the results of cheap operations such as batch normalization \cite{ioffe2015batch} or ReLU can be dropped and then recomputed later.
All the gradient check-pointing approaches add processing overhead during training.
\par
{\bf Semantic Segmentation.}
Current state-of-the-art methods for semantic segmentation are based on convolutional neural networks.
These networks have different architectures.
Encoder-decoder or hourglass networks are used in many computer vision tasks like object detection \cite{lin2017feature}, human pose estimation \cite{newell2016stacked}, image-based localization \cite{melekhov2017image}, and semantic segmentation \cite{long2015fully,badrinarayanan2017segnet,noh2015learning}. Generally, they are made of an encoder and decoder parts such that, the encoder gradually reduces the feature maps resolution and captures high-level semantic information, and the decoder gradually recovers the low-level details.
Because these networks lose the image details during the encoder path, they are not able to achieve the highest results without using skip connections.
Spatial pyramid pooling models perform spatial pyramid pooling \cite{lazebnik2006beyond,grauman2005pyramid} at different grid scales or apply several parallel atrous convolution \cite{chen2017deeplab} with different rates. These models include the two well-known PSPNet \cite{zhao2017pyramid} and DeepLab \cite{chen2017rethinking}.
High-resolution representation networks \cite{wang2019deep,huang2017multi,fourure2017residual,zhou2015interlinked} try to maintain a high-resolution hidden state from input to output. By doing low-resolution convolutions in parallel streams, high-level features are gained while low-level details are not lost.
Since these networks require a lot of memory, they first down-sample the input image to a lower resolution before the main body.
Some approaches \cite{chen2017deeplab,chandra2016fast} do post-processing, such as conditional random fields, on the network's output to improve the segmentation details, especially around the object boundaries.
These approaches add some processing overhead to training and testing.
\par
{\bf Semantic Segmentation using Neural ODEs.} 
There are only a few methods for semantic segmentation that have partially incorporated neural ODEs in the network design.
In \cite{pinckaers2019neural}, a U-Net is modified to use neural ODEs.
In this design, the repeated residual blocks in each branch are replaced by a neural ODE that wraps around only one convolutional block.
Although U-Net is a well-known network, more recent networks can achieve higher results than U-Net.
In this paper, we design our network based on the HRNetV2 \cite{wang2019deep} which can achieve the state-of-the-art accuracy on the Cityscapes \cite{cordts2016cityscapes}, CamVid \cite{BrostowFC:PRL2008}, and LIP \cite{gong2017look} datasets.
Similar to \cite{pinckaers2019neural}, another modified U-Net is introduced in \cite{li2021robust}.
This time, instead of replacing a branch with a neural ODE block, a neural ODE block is added at the end of each branch.
In \cite{valle2019neural}, a novel approach that combines neural ODEs and the Level Set method is proposed.
This approach parameterizes the derivative of the contour as a neural ODE that implicitly learns a forcing function describing the evolution of the contour.
This approach is limited to the segmentation of images with one target class.

\begin{figure}[t]
\includegraphics[width=\textwidth]{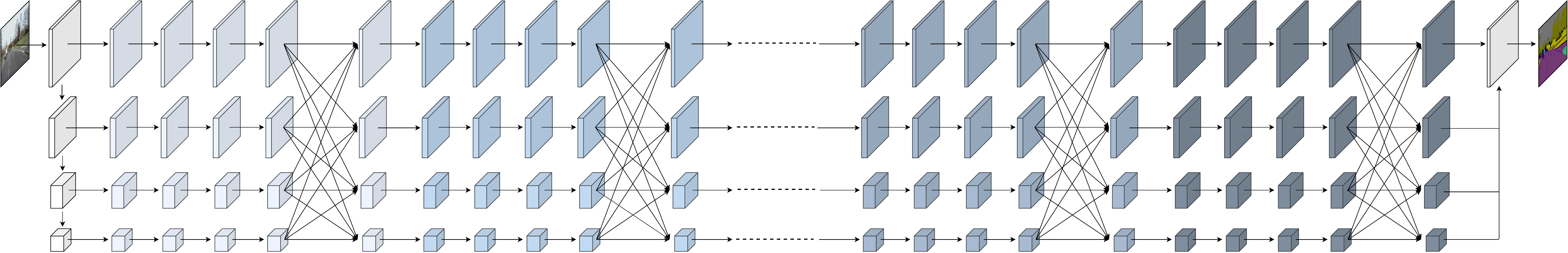}
\caption{The baseline network is created by repeating the last module from HRNetV2 \cite{wang2019deep} which has four branches with different feature-map resolutions.
We use skip connections at the module level (not drawn).
So, each module is treated as a residual block.
Each small block in a module consists of one set of batch normalization \cite{ioffe2015batch}, ReLU, and convolutional layers.
This network is not a neural ODE and is trained similarly to HRNetV2.}
\label{fig:base-network}
\end{figure}

\section{Method}
We introduce a baseline network that is trained without the use of neural ODEs.
Then we introduce a neural ODE equivalent to the baseline network.
Our goal is to compare the results step by step, from a state-of-the-art network to the baseline network, then to the neural ODE network.

\subsection{Baseline Network}
At the time of writing, one of the state-of-the-art methods in semantic segmentation is HRNetV2 \cite{wang2019deep}.
We try to adopt this network architecture and turn it into a residual form such that each module is like a residual block.
To this aim, we repeat the last module in series, multiple times and treat each one of them as a residual unit (as depicted in Figure \ref{fig:base-network}).
This way, the network consists of multiple residual modules, each module has four branches with different feature-map resolutions, and each branch has multiple residual blocks.
In our experiments in this paper, we repeat the main module six times to keep the number of parameters close to HRNetV2.
We use this baseline network to gradually evaluate the design of our neural ODE network.

\subsection{SegNode}
Since the baseline network has an overall residual form, we can turn it into a neural ODE.
In this form, a single or multiple modules act as the function $f$ in Equation \ref{eq:1}.
This module (or modules) is wrapped in an ODE solver.
Since the main module has four convolutional streams with different resolutions and number of channels, we use convolutional layers to create the input feature-maps with the corresponding resolution and number of channels.
The resulting four tensors are fed to the ODE solver.
The output of the ODE solver has the same format as its input.
By using four convolutional layers, the number of channels of the output feature maps is changed to the number of classes.
Then, the feature maps are re-scaled to the higher resolution using bilinear interpolation and added together to produce the final output (as shown in Figure \ref{fig:segnode}).
Bilinear interpolation is used to avoid the checkerboard artifact \cite{odena2016deconvolution}.
We call our network SegNode for short.

\begin{table}[t]
\caption{Comparison of results on four datasets.
We use \dag ~to mark methods pretrained on Mapillary.}
\label{tab:results}
\centering
\setlength\tabcolsep{8pt}
\begin{tabular}{l|cccc}
 \hline
  Method & Cityscapes & CamVid & LIP & PASCAL-Context \\
 \hline
 HRNetV2 \cite{wang2019deep} & 81.6 & 80.9 & 55.9 & 54.0 \\
 HRNetV2+OCR \cite{yuan2020object} & 83.0 & 81.7 & 56.6 & 56.2 \\
 HRNetV2+OCR\textsuperscript{\dag} \cite{yuan2020object} & 84.2 & - & - & - \\
 \hline
 U-Node \cite{pinckaers2019neural} & 78.1 & 77.3 & 51.3 & 49.7 \\
 NODEs-UNet \cite{li2021robust} & 79.5 & 78.8 & 52.9 & 50.9 \\
 \hline
 Baseline network & 81.7 & 81.0 & 55.9 & 53.9 \\
 SegNode & 81.8 & 81.1 & 55.8 & 54.1 \\
 SegNode+OCR & 83.1 & {\bf 82.0} & {\bf 56.7} & {\bf 56.2} \\
 SegNode+OCR\textsuperscript{\dag} & {\bf 84.5} & - & - & - \\
 \hline
\end{tabular}
\end{table}

\section{Experiments}
We evaluate our approach on four datasets: Cityscapes \cite{cordts2016cityscapes}, CamVid \cite{BrostowFC:PRL2008}, LIP \cite{gong2017look}, and PASCAL-Context \cite{mottaghi2014role}.
Additionally, since the existing neural ODE methods for semantic segmentation have not been evaluated on these datasets, we train and test the two U-Net based methods \cite{pinckaers2019neural,li2021robust} on these datasets and report their accuracy.

\subsection{Setup}
We pretrain our baseline and SegNode networks on ImageNet \cite{russakovsky2015imagenet} and use the pre-trained networks in all our experiments.
We use the mean Intersection over Union (mIoU) metric to compare all the methods.
\par
For the baseline network, we use AdamW optimizer \cite{loshchilov2017decoupled} with a weight decay of 0.05 and batch size of 16.
We apply the "polynomial" learning rate policy with a poly exponent of 0.9 and an initial learning rate of 0.0001.
\par
For SegNode, we use the Runge-Kutta ODE solver provided by \cite{chen2018neural}.
Also, we use the adjoint sensitivity method \cite{pontryagin2018mathematical} which is available in the same implementation.
We use the SGD optimizer with a base learning rate of 0.1, a momentum of 0.9, and no weight decay.
The polynomial learning rate decay function is used with a poly exponent of 0.9.
\par
For both the baseline network and SegNode, similar to HRNetV2 \cite{wang2019deep}, we use a stem for the input image, which consists of two stride-2 3$\times$3 convolutions to decrease the resolution to $1/4$, and is connected to the main body.
The main body outputs the feature maps with the same resolution ($1/4$), which are then made larger as the original resolution using bilinear interpolation.
Each stream in the main body has 48, 96, 192, and 384 channels respectively from the highest resolution to the lowest.
We use two modules in the main body to achieve the highest accuracy.

\begin{table}[t]
\begin{center}
\setlength\tabcolsep{4pt}
\begin{tabular}{l|cccccc}
  Method
  & \rotatebox{90}{Maximum Batch Size for 32GB}
  & \rotatebox{90}{Memory Usage for Batch Size 24 (GB)}
  & \rotatebox{90}{Memory Usage for Testing One Image (GB)}
  & \rotatebox{90}{Training Time per Epoch (Seconds)}
  & \rotatebox{90}{Testing Time per Image (Milliseconds)}
  & \rotatebox{90}{Number of Parameters (Millions)} \\
 \hline
 U-Net \cite{ronneberger2015u} & 36 & 21.8 & 0.8 & {\bf 10} & {\bf 4} & 31.0 \\
 PSPNet \cite{zhao2017pyramid} & 24 & 31.2 & 0.8 & 18 & 5 & 23.7 \\
 Deeplab v3 \cite{chen2017rethinking} & 24 & 31.3 & 1.0 & 24 & 16 & 58.6 \\
 HRNetV2 \cite{wang2019deep} & 24 & 31.1 & 1.2 & 18 & 48 & 65.8 \\
 \hline
 Baseline network & 24 & 31.9 & 1.2 & 19 & 49 & 70.9 \\
 SegNode & {\bf 62} & {\bf 13.4} & {\bf 0.7} & 34 & 117 & {\bf 20.9} \\
 \hline
\end{tabular}
\end{center}
\caption{A comparison of a few important empirical computational measures on an NVIDIA Tesla V100 32GB for CamVid \cite{BrostowFC:PRL2008} dataset.
The training time per epoch is calculated using the maximum batch size possible.
Our method requires the least amount of memory, but the longest computation time for training and testing.}
\label{tab:computational-cost}
\end{table}

\subsection{Cityscapes}
The Cityscapes dataset \cite{cordts2016cityscapes} contains 5k high quality pixel-level finely annotated street images.
The finely annotated images are divided into 2,975/500/1,525 images for training, validation, and testing.
Also, the dataset contains additional 20k coarsely annotated images.
There are 30 classes, and 19 classes among them are used for evaluation.
We train on the training, validation, and coarse sets to get the highest accuracy on the test set.

\subsection{CamVid}
Compared to Cityscapes \cite{cordts2016cityscapes}, CamVid \cite{BrostowFC:PRL2008} is a much smaller dataset focusing on semantic segmentation for driving scenarios.
The original version is composed of 701 annotated images in 32 classes with size 960$\times$720 from five video sequences.
However, most literature only focuses on the protocol proposed in \cite{badrinarayanan2017segnet} which splits the dataset into 367 training, 101 validation, and 233 test images in 11 classes.
We follow this protocol for training on CamVid.

\subsection{LIP}
The LIP dataset \cite{gong2017look} contains 50,462 human images with detailed annotations.
The dataset is divided into 30,462 training, 10,000 validation, and 10,000 test images.
The model evaluation is done on 20 categories (including the background label).
We follow the common testing protocol \cite{ruan2019devil,wang2019deep} and resize the images to 473$\times$473.

\subsection{PASCAL-Context}
The PASCAL-Context dataset \cite{mottaghi2014role} adds annotations for more than 400 additional categories to the PASCAL VOC 2010 dataset.
It contains 4,998 training and 5,105 validation images, subsets of PASCAL VOC 2010 dataset.
The dataset annotations cover 100\% of pixels while the previous annotations covered around 29\%.
We follow \cite{wang2019deep,yuan2020object} and evaluate our method on 59 sub-categories.

\begin{figure}[t]
\includegraphics[width=\textwidth]{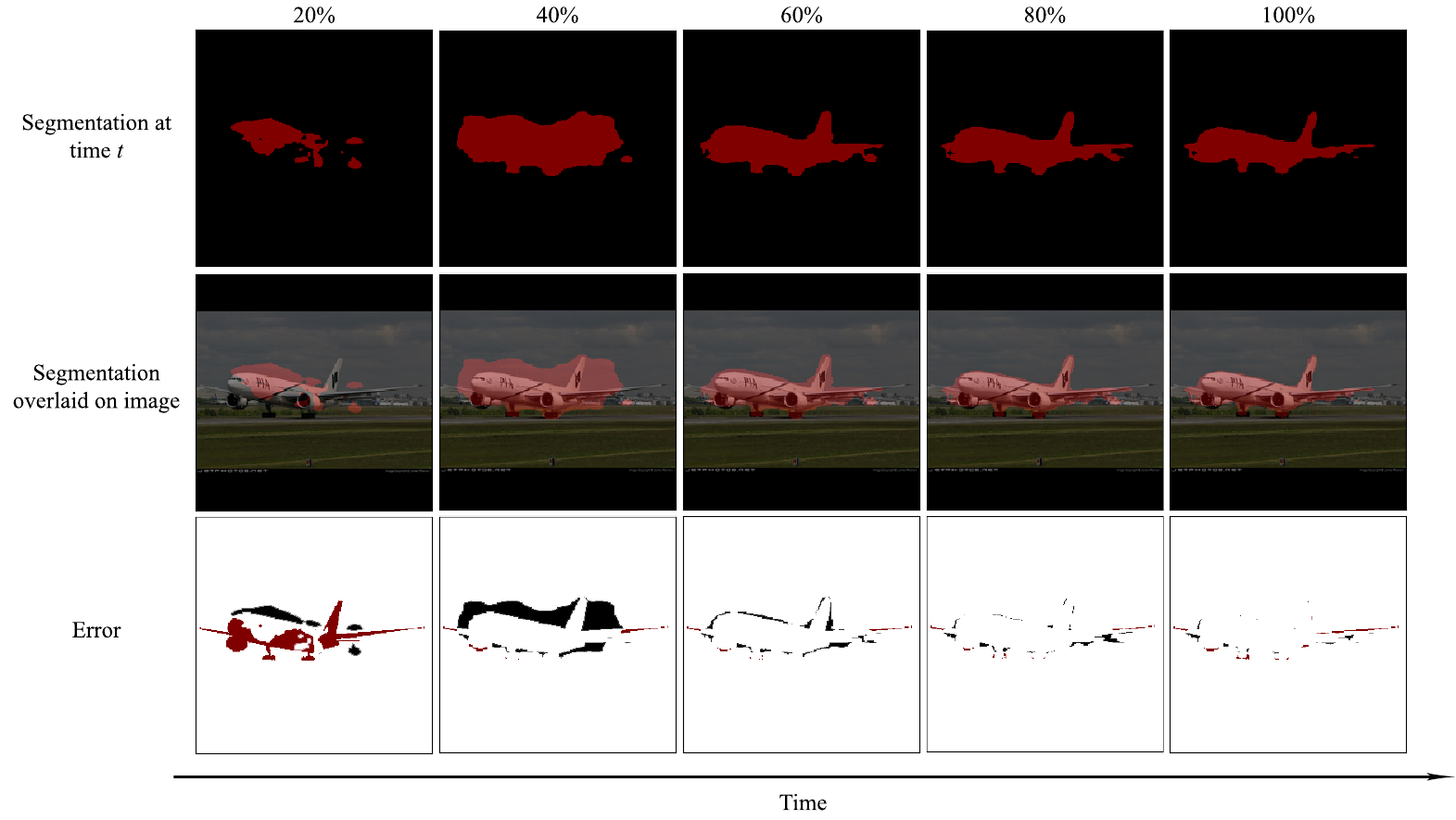}
\caption{Segmentation results from trajectories at different times.
This image shows how the gradual transformations correct the segmentation over time.}
\label{fig:trajectories}
\end{figure}

\subsection{Results}
Table \ref{tab:results} compares the results of our proposed method to different variants of HRNetV2 and existing neural ODE methods.
On average our method performs better than HRNetV2 and its variants by a small margin.
\par
We tried and improved the existing neural ODE methods in our implementation by increasing the number of parameters and tuning hyper-parameters.
Still, our proposed design can achieve higher accuracy by a large margin.
The main reason is that we started our design from a better-performing network architecture and modified it step-by-step towards the final design.

\begin{figure}[t]
\centering
\includegraphics[width=0.8\textwidth]{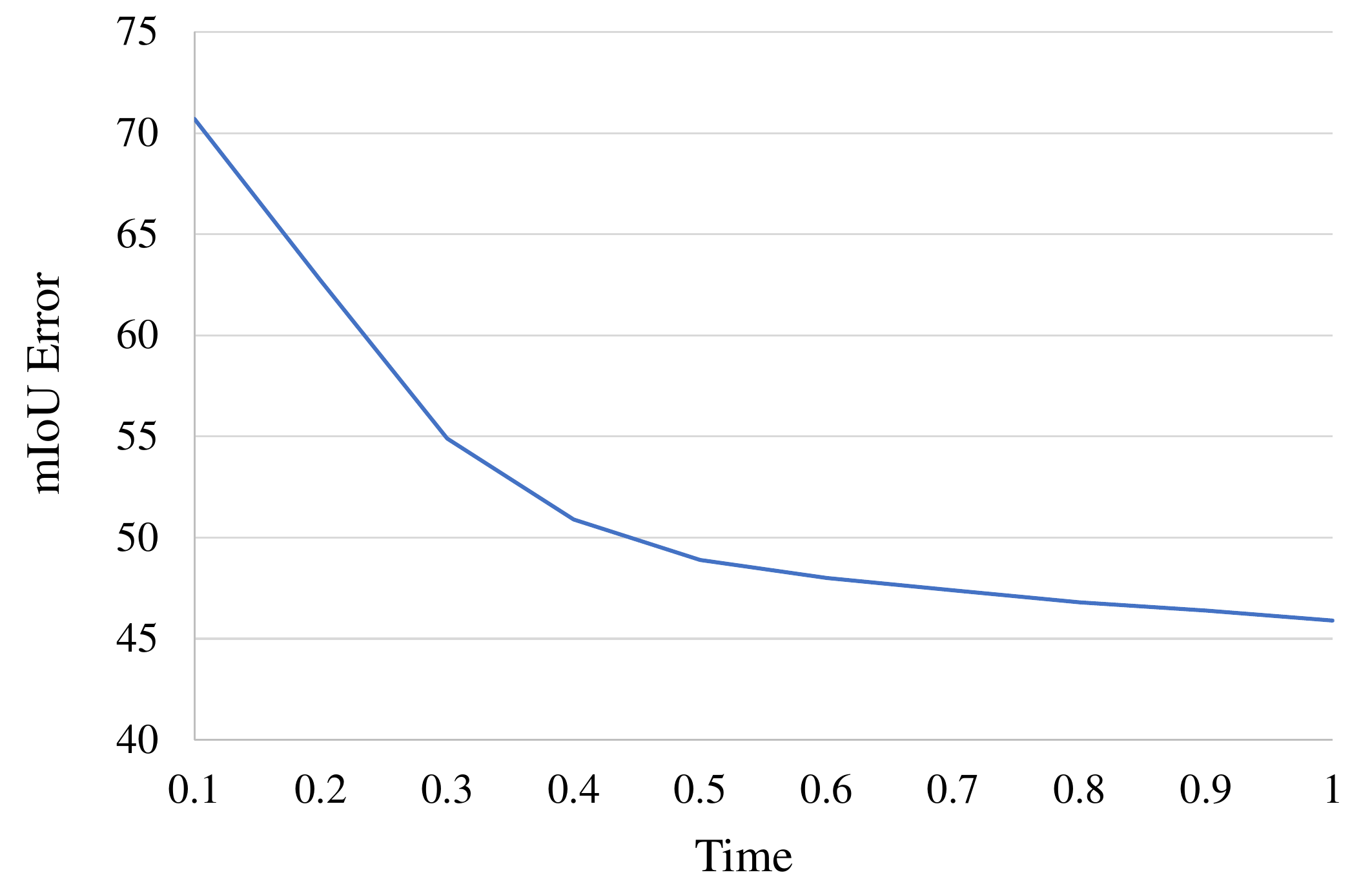}
\caption{The average mean IoU error of the trajectories during solving time, calculated on the PASCAL-Context validation set.}
\label{fig:error}
\end{figure}

\subsection{Empirical Computational Cost} \label{computational-cost}
In this section, we provide an empirical comparison between our approach and a few well-known networks.
We use an NVIDIA Tesla V100 32GB with the same network hyper-parameters as used before.
All the experiments are implemented in Python using PyTorch.
The results are calculated on CamVid \cite{BrostowFC:PRL2008} dataset with an image size of $480\times360$.
\par
Table \ref{tab:computational-cost} compares a few important empirical computational measures.
Our method requires the least amount of memory, but the longest computation time for training and testing by a large margin.
In particular, compared to HRNetV2 \cite{wang2019deep}, while our method has 68\% less number of parameters, it requires 57\% less memory for training and 42\% less memory for testing.
On the other side, HRNetV2 requires 47\% less training time and 59\% less testing time.

\subsection{Trajectory Error}
In this section, we show how the ODE solver gradually improves its output during test time.
Figure \ref{fig:trajectories} visualizes the segmentation output of the network trajectories over time for one sample image.
This figure shows the steps that the ODE solver takes during solving the network.
To generate each step, the corresponding hyper-parameter of the ODE solver is modified to partially solve its input.
\par
Figure \ref{fig:error} shows the average mean IoU error of the trajectories over time for all the samples in the PASCAL-Context validation set.
One of the biggest issues with neural ODEs is that they require more computational resources during test time.
To alleviate this problem, it is possible to sacrifice accuracy for speed.
As an example, by sacrificing 3\% of accuracy, the required computational time decreases by 50\%.

\section{Conclusion}
Based on a current state-of-the-art network, we proposed a novel neural ODE design for semantic segmentation.
The new idea of neural ODEs helped us to reduce the memory requirement with the cost of more processing time.
While using a notably less amount of memory, our method (SegNode) was able to achieve state-of-the-art results.
The proposed method can be used for all the computer vision tasks that can make use of dense 2D predictions such as human pose estimation and object detection tasks.

\bibliographystyle{splncs04}
\bibliography{references}

\end{document}